\documentclass{article}
\usepackage{spconf-joan}

\usepackage{graphicx}
\usepackage{amsmath,amssymb,bm}
\usepackage{url}
\usepackage{setspace}
\usepackage[noadjust]{cite}
\usepackage{multicol}
\usepackage{flushend}

\usepackage{hyperref}
\usepackage[dvipsnames]{xcolor}
\newcommand\myshade{70}
\colorlet{mywholecolor}{MidnightBlue}
\hypersetup{
  linkcolor  = mywholecolor!\myshade!black,
  citecolor  = mywholecolor!\myshade!black,
  urlcolor   = mywholecolor!\myshade!black,
  colorlinks = true,
}

\newcommand{\ve}[1]{\textbf{#1}}

\usepackage{color}

\newcommand\sectitle[1]{\vspace{0.4cm}\begin{center}\uppercase{\textbf{#1}}\end{center}}

\usepackage[english]{babel}
\title{On tuning consistent annealed sampling for denoising score matching}
\name{Joan Serr\`a, Santiago Pascual, Jordi Pons}
\address{Dolby Laboratories}
\begin{document}
\onecolumn
%
\maketitle
\begin{abstract}
Score-based generative models provide state-of-the-art quality for image and audio synthesis. Sampling from these models is performed iteratively, typically employing a discretized series of noise levels and a predefined scheme. In this note, we first overview three common sampling schemes for models trained with denoising score matching. Next, we focus on one of them, consistent annealed sampling, and study its hyper-parameter boundaries. We then highlight a possible formulation of such hyper-parameter that explicitly considers those boundaries and facilitates tuning when using few or a variable number of steps. Finally, we highlight some connections of the formulation with other sampling schemes.
\end{abstract}


\vspace{5pt}
\begin{multicols}{2}


\sectitle{Introduction}

\noindent
Score matching~\cite{hyvarinen_estimation_2005} has become a successful approach to train energy-based models~\cite{song_how_2021}. Score-based generative models have recently demonstrated state-of-the-art quality for image~\cite{ho_denoising_2020, song_score-based_2021} and audio~\cite{chen_wavegrad_2021, kong_diffwave_2021} synthesis. Such models are typically trained with a denoising score matching objective~\cite{vincent_connection_2011} to approximate the score of the data distribution $p(\ve{x})$, $s_\theta \approx \nabla_{\ve{x}}\log p(\ve{x})$. To alleviate discontinuities in $p(\ve{x})$ and facilitate differentiation, data samples can be perturbed using geometrically-spaced noise levels $\sigma_1 > \sigma_2 > \dots > \sigma_N$, $\gamma=\sigma_{i+1}/\sigma_i$~\cite{song_generative_2019}, where $\sigma_1$ is large enough to cover $p(\ve{x})$ and $\sigma_N$ is small enough to be close to imperceptible~\cite{song_improved_2020}. 

Synthesis is typically performed with annealed Langevin sampling (ALS)~\cite{song_generative_2019}, which recursively iterates from $\sigma_1$ to $\sigma_N$. Nonetheless, a continuum between $\sigma_1$ and $\sigma_N$, $\sigma_i\in[\sigma_1,\sigma_N]$, may also be employed at training time~\cite{chen_wavegrad_2021}. This is useful in practice, as the same model can be sampled under different computational budgets (different $N$) without the need of retraining. Furthermore, training with a continuum opens up the possibility of fine-tuning individual $\sigma_i$ for sampling, with the potential to achieve convincing results when using a small amount of iterations (see~\cite{chen_wavegrad_2021}).

Given a sufficiently accurate estimation of the score $s_\theta$, ALS recursively iterates across noise levels $\sigma_i$, $i=1,\dots N$, following
\begin{equation}
\ve{x}_i \leftarrow \ve{x}_{i-1} + \alpha s_\theta(\ve{x}_{i-1},\sigma_i) + \sqrt{2\alpha}\ve{z}_i,
\label{eq:als}
\end{equation}
where $\alpha=\epsilon_a\sigma_i/\sigma_N$ controls the step size ($\epsilon_a$ is an hyper-parameter) and $\ve{z}_i \sim \mathcal{N}(0,\ve{I})$. Since $\sigma_1$ is usually much larger than the standard deviation of the signal of interest, one can ignore the latter at the beginning of the process and use $\ve{x}_0 \sim \mathcal{N}(0,\sigma_1\ve{I})$.

In recent work, Jolicoeur-Martineau et al.~\cite{jolicoeur-martineau_adversarial_2021} point out some inconsistencies for ALS in the context of score-based generative models. In particular, given that $N\ll\infty$ and that $s_\theta$ is just an approximation of the true score function, they show that the noise level in ALS does not follow the prescribed schedule. To tackle this, they propose consistent annealed sampling (CAS), which employs the recursion
\begin{equation}
\ve{x}_i \leftarrow \ve{x}_{i-1} + \eta \sigma_i^2 s_\theta(\ve{x}_{i-1},\sigma_i) + \beta\sigma_{i+1}\ve{z}_i,
\label{eq:cas}
\end{equation}
where $\eta=\epsilon_b/\sigma_N^2$ controls the step size ($\epsilon_b$ is an hyper-parameter) and
\begin{equation}
\beta = \sqrt{1 - \left( \frac{1-\eta}{\gamma} \right)^2 } .
\label{eq:beta}
\end{equation}
The same authors also demonstrate better synthesis quality by taking the expected denoised sample $H(\ve{x},\sigma)=\ve{x}+\sigma^2s_\theta(\ve{x},\sigma)$ at the last iteration~\cite{jolicoeur-martineau_adversarial_2021}:
\begin{equation}
\ve{x}_{N} \leftarrow \ve{x}_{N-1} + \sigma_N^2 s_\theta(\ve{x}_{N-1},\sigma_N) .
\label{eq:eds}
\end{equation}

In work concurrent to CAS, Song et al.~\cite{song_score-based_2021} propose to sample through a predictor-corrector (PC) scheme, which is based on discretized, reverse-time stochastic differential equations (SDEs). PC schemes sample by combining an outer-loop prediction stage that can be written as
\begin{equation}
\ve{x}_i \leftarrow \ve{x}_{i-1} + (\sigma_{i}^2-\sigma_{i+1}^2) s_\theta(\ve{x}_{i-1},\sigma_{i}) + \sqrt{\sigma_{i}^2-\sigma_{i+1}^2} \ve{z}_i
\label{eq:pc}
\end{equation}
with an inner-loop correction stage based on ALS such that, after every step with Eq.~\ref{eq:pc}, $i=1,\dots N$, $M$ steps are performed with Eq.~\ref{eq:als}. This PC scheme corresponds to what the authors call the ``variance exploding'' SDE, which matches the case considered by CAS and previous works using ALS~\cite{song_generative_2019, song_improved_2020}. Note that, because they use ALS, PC schemes can also present inconsistencies for small $N$ and/or a suboptimal $s_\theta$.


\begin{figure*}[tb]
\center
\includegraphics{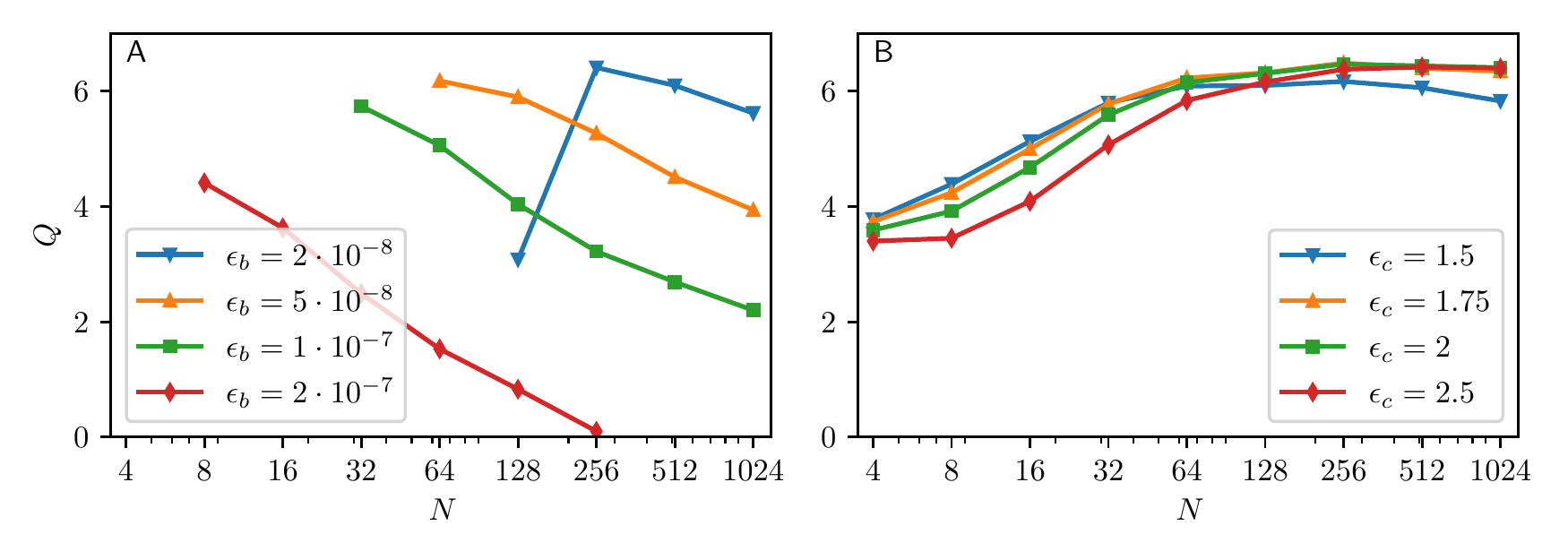}
\vspace{-20pt}
\caption{Examples of performance curves when tuning $\epsilon_b$ (A) and $\epsilon_c$ (B) for different number of sampling steps $N$. We use a model of speech synthesis, and the performance measure $Q$ corresponds to a composite quality measure featuring SESQA~\cite{serra_sesqa_2021} (the higher the better).}
\label{fig}
\end{figure*}

\sectitle{Tuning consistent annealed sampling}

\noindent
We are interested in performing synthesis with as few steps as possible, and studying synthesis quality as a function of $N$ and $\epsilon$. As mentioned, if we want a model to be able to work under different computational budgets, we can train it with a continuum of $\sigma_i\in[\sigma_1,\sigma_N]$. Because we want to avoid inconsistencies with small $N$, we choose CAS as our sampling scheme and tune $\epsilon_b$. 
To do so, however, it is advisable to keep in mind the boundaries that make sense for $\eta$, as $\eta=\epsilon_b/\sigma_N^2$. In particular, from Eq.~\ref{eq:beta}, we have that
\begin{equation*}
\left( \frac{1-\eta}{\gamma} \right)^2 \le 1 ,
\end{equation*}
otherwise $\beta$ becomes an imaginary number. This implies that $1-\gamma\le\eta\le 1+\gamma$. To further tighten the upper bound, we can additionally have a look at Eq.~\ref{eq:cas}, where we see that, unless $\eta\le1$, we risk of amplifying the noise at every iteration~\cite{jolicoeur-martineau_adversarial_2021} (compare also with Eq.~\ref{eq:eds}). Thus, taking the two constraints together, we have
\begin{equation}
1-\gamma \le \eta \le 1 .
\label{eq:boundaries}
\end{equation}

It is intuitive to think from the previous boundaries that $\eta$, and therefore $\epsilon_b$, should have a direct relation with $\gamma$, which in turn depends on $N$. Thus, one could conclude that the hyper-parameter $\epsilon_b$ is tied to $N$, and that the best value for the former will strongly depend on the latter (we empirically verify it below). This is a problem for a model/sampling that aims to operate at different computational budgets, or simply for studying how quality depends on our choices of $N$ and $\epsilon_b$.

To circumvent this problem, we propose to parameterize $\eta$ by taking into account the effect that different $N$ have into the geometric progression of the noise levels $\sigma$ that will be used for sampling. More specifically, we propose to use
\begin{equation}
\eta = 1-\gamma^{\epsilon_c} ,
\label{eq:neweta}
\end{equation}
where $\epsilon_c\ge1$ is the new version of the sampling hyper-parameter. This expression naturally encapsulates the strength of the relation of $N$ with $\eta$ through the ratio of the geometric progression $\gamma$. The hyper-parameter $\epsilon_c$ then acts as a tunable modifier of such strength, with $\epsilon_c=1$ and $\epsilon_c\rightarrow\infty$ yielding the lower and upper boundaries of Eq.~\ref{eq:boundaries}, respectively (recall that $0<\gamma<1$ if one takes a geometric progression).

The benefits of the formulation in Eq.~\ref{eq:neweta} are illustrated in Fig.~\ref{fig}. If we manually search for the best hyper-parameter following the original formulation, $\eta=\epsilon_b/\sigma_N^2$, we observe that the best $\epsilon_b$ is hard to find and rapidly changes with $N$ (Fig.~\ref{fig}A). We are able to find some good operation points at different $N$, but these are only found for disparate values of $\epsilon_b$ (notice the order of magnitude difference between $N=8$ and $N=256$). Not only that, but the sampling breaks for small $N$ and it is not very stable for increasing $N$, which is due to $\eta$ hitting or getting close to the boundary conditions as we modify $N$ (and therefore $\gamma$). Contrastingly, if we manually search for the best hyper-parameter following the proposed formulation, $\eta = 1-\gamma^{\epsilon_c}$, we observe that the best $\epsilon_c$ is easier to find and remains more stable with $N$ (Fig.~\ref{fig}B). Good operation points are found across a wide range of $N$ and with multiple values of $\epsilon_c$. Importantly, sampling now does not break for small $N$, as $\eta$ remains properly bounded.


\sectitle{Relation to other sampling schemes}

\noindent
Given the formulation of $\eta$ in Eq.~\ref{eq:neweta}, we find that a particular value of $\epsilon_c$ allows to compare CAS with other sampling schemes. Specifically, if we choose $\epsilon_c=2$, we have $\eta=1-\gamma^2$ and, by substitution in Eq.~\ref{eq:beta}, $\beta=\sqrt{1-\gamma^2}=\sqrt{\eta}$. This allows to write the CAS recursion (Eq.~\ref{eq:cas}) as
\begin{equation*}
\ve{x}_i \leftarrow \ve{x}_{i-1} + \eta\sigma_i^2 s_\theta(\ve{x}_{i-1},\sigma_i) + \sqrt{\eta\sigma_{i+1}^2}\ve{z}_i
\end{equation*}
which, using $\gamma=\sigma_{i+1}/\sigma_i$, becomes
\begin{equation}
\ve{x}_i \leftarrow \ve{x}_{i-1} + \eta\sigma_i^2 s_\theta(\ve{x}_{i-1},\sigma_i) + \gamma\sqrt{\eta\sigma_{i}^2}\ve{z}_i .
\label{eq:relat}
\end{equation}
Notice that this version of CAS, obtained by choosing $\epsilon_c=2$, is similar to ALS. In particular, if we assume $\alpha'=\eta\sigma_i^2$, we have 
\begin{equation*}
\ve{x}_i \leftarrow \ve{x}_{i-1} + \alpha' s_\theta(\ve{x}_{i-1},\sigma_i) + \gamma\sqrt{\alpha'}\ve{z}_i ,
\end{equation*}
which has the same form as Eq.~\ref{eq:als}, except for a factor of $\gamma/\sqrt{2}$ in the noise term. 

Notice also that the same version of CAS using $\epsilon_c=2$ can be related to the predictor part of the PC scheme. In particular, if we substitute $\eta=1-\gamma^2$ in Eq.~\ref{eq:relat} and operate, we obtain
\begin{equation*}
\ve{x}_i \leftarrow \ve{x}_{i-1} + (\sigma_i^2-\gamma^2\sigma_i^2) s_\theta(\ve{x}_{i-1},\sigma_i) + \gamma\sqrt{\sigma_{i}^2-\gamma^2\sigma_{i}^2}\ve{z}_i 
\end{equation*}
which, using again $\gamma=\sigma_{i+1}/\sigma_i$, becomes
\begin{equation*}
\ve{x}_i \leftarrow \ve{x}_{i-1} + (\sigma_i^2-\sigma_{i+1}^2) s_\theta(\ve{x}_{i-1},\sigma_i) + \gamma\sqrt{\sigma_{i}^2-\sigma_{i+1}^2}\ve{z}_i .
\end{equation*}
This is the same expression as Eq.~\ref{eq:pc}, except for a factor of $\gamma$ in the noise term. Interestingly, this shows that, for the case of $\epsilon_c=2$ and $\gamma=1$ ($N\rightarrow\infty$), CAS becomes the predictor part of the PC scheme. It also suggests that the predictor part of the PC scheme is consistent for $\gamma=1$ (and therefore for $N\rightarrow\infty$).

Apart from $\epsilon_c=2$, we can also consider the cases of $\epsilon_c=1$ and $\epsilon_c\rightarrow\infty$, corresponding to the boundaries outlined in the previous section. Consider again Eqs.~\ref{eq:cas} and~\ref{eq:beta}. For $\epsilon_c=1$, we have $\eta=1-\gamma$ and $\beta=0$, which yields a recursion without including any noise component beyond $\ve{x}_0$:
\begin{equation*}
\ve{x}_i \leftarrow \ve{x}_{i-1} + (1-\gamma) \sigma_i^2 s_\theta(\ve{x}_{i-1},\sigma_i) .
\end{equation*}
This corresponds to a recursion solely based on empirically denoising partial samples. We see that by considering $H(\ve{x},\sigma)=\ve{x}+\sigma^2s_\theta(\ve{x},\sigma)$ and operating, which yields:
\begin{equation*}
\ve{x}_i \leftarrow \gamma\ve{x}_{i-1} + (1-\gamma)H(\ve{x}_{i-1},\sigma_i) .
\end{equation*}
For $\epsilon_c\rightarrow\infty$, we have $\eta=1$ and $\beta=1$, which yields 
\begin{equation*}
\ve{x}_i \leftarrow \ve{x}_{i-1} + \sigma_i^2 s_\theta(\ve{x}_{i-1},\sigma_i) + \sigma_{i+1}\ve{z}_i .
\end{equation*}
This corresponds to a recursion solely based on adding noise to the empirically denoised sample. The first two terms directly correspond to it:
\begin{equation*}
\ve{x}_i \leftarrow H(\ve{x}_{i-1},\sigma_i) + \sigma_{i+1}\ve{z}_i .
\end{equation*}
The hyper-parameter $\epsilon_c\in[1,\infty)$ allows to smoothly interpolate between all cases considered here.


\end{multicols}

\vspace{3pt}
\begin{center}
--------------------------------
\end{center}
\vspace{2pt}

\begin{multicols}{2}

\bibliographystyle{IEEEbib}
\bibliography{main}

\end{multicols}

\end{document}